\documentclass[runningheads]{llncs}

\usepackage{eccv}

\usepackage{eccvabbrv}
\usepackage{xspace}
\usepackage{graphicx}
\usepackage{booktabs}

\usepackage[accsupp]{axessibility}  % Improves PDF readability for those with disabilities.

\usepackage[pagebackref,breaklinks,colorlinks,citecolor=eccvblue]{hyperref}
\usepackage{orcidlink}

\usepackage{booktabs}
\usepackage{graphicx}
\usepackage{amsmath}
\usepackage{amssymb}
\usepackage{multirow}
\usepackage{multicol}
\usepackage{adjustbox} 
\usepackage{graphicx, lipsum}
\usepackage{wrapfig}
\usepackage{pifont} 
\usepackage{bm} 
\usepackage[font=normalsize,labelfont=bf,tableposition=top]{caption}
\usepackage{subcaption}
\usepackage{array}
\newcommand{\cmark}{\ding{51}}

\begin{document}


\title{WorldAfford: Affordance Grounding based on Natural Language Instructions} 

\titlerunning{Affordance Grouding}

\author{Changmao Chen\inst{1}, Yuren Cong\inst{2}, Zhen Kan\inst{1}}

\authorrunning{Chen et al.}

\institute{University of Science and Technology of China \and
TNT, Leibniz University Hannover
\\
\email{\{abc,lncs\}@uni-heidelberg.de, cong@tnt.uni-hannover.de}}

\maketitle
\textbf{}

%Abstract
% 1. The definition of the proposed task. 
% 2. The main challenge of the task.
% 3. Drawbacks of the previous works.
% 4. A high-level summary of our approach
% 5. Experiment results show our method has superior performance...
\begin{abstract}

Affordance grounding aims to localize the interaction regions for the manipulated objects in the scene image according to given instructions, which is essential for Embodied AI and manipulation tasks. 
A critical challenge in affordance grounding is that the embodied agent should understand human instructions and analyze which tools in the environment can be used, as well as how to use these tools to accomplish the instructions.
Most recent works primarily supports simple action labels as input instructions for localizing affordance regions, failing to capture complex human objectives. Moreover, these approaches typically identify affordance regions of only a single object in object-centric images, ignoring the object context and struggling to localize affordance regions of multiple objects in complex scenes for practical applications.
To address this concern, for the first time, we introduce a new task of affordance grounding based on natural language instructions, extending it from previously using simple labels for complex human instructions.
For this new task, we propose a new framework, \textbf{WorldAfford}.
We design a novel Affordance Reasoning Chain-of-Thought Prompting to reason about affordance knowledge from LLMs more precisely and logically.
Subsequently, we use SAM and CLIP to localize the objects related to the affordance knowledge in the image. 
We identify the affordance regions of the objects through an affordance region localization module.
To benchmark this new task and validate our framework, an affordance grounding dataset, LLMaFF, is constructed.
We conduct extensive experiments to verify that WorldAfford performs state-of-the-art on both the previous AGD20K and the new LLMaFF dataset.
In particular, WorldAfford can localize the affordance regions of multiple objects and provide an alternative when objects in the environment cannot fully match the given instruction.
The code will be released after the publication of this work.

\end{abstract}

\section{Introduction}
\label{sec:intro}
Embodied agents can interact with a physical environment and potentially perform heavy tasks based on human instructions.  
In order for robots to better manipulate objects in complex scenes, it is urgent to understand which part of the object is the interaction region.
Affordance grounding, which aims to localize potential interaction regions for the manipulated objects in the scene image depending on the given instruction, can provide a new experience for Embodied AI and has the potential to significantly increase efficiency and flexibility. 
As a result, it has recently attracted a significant amount of attention~\cite{hassanin2021visual, luo2022learning, zhai2022one, li2023locate, luo2023grounded}.

\begin{figure}[ht]
\centering
\includegraphics[width=1\linewidth]{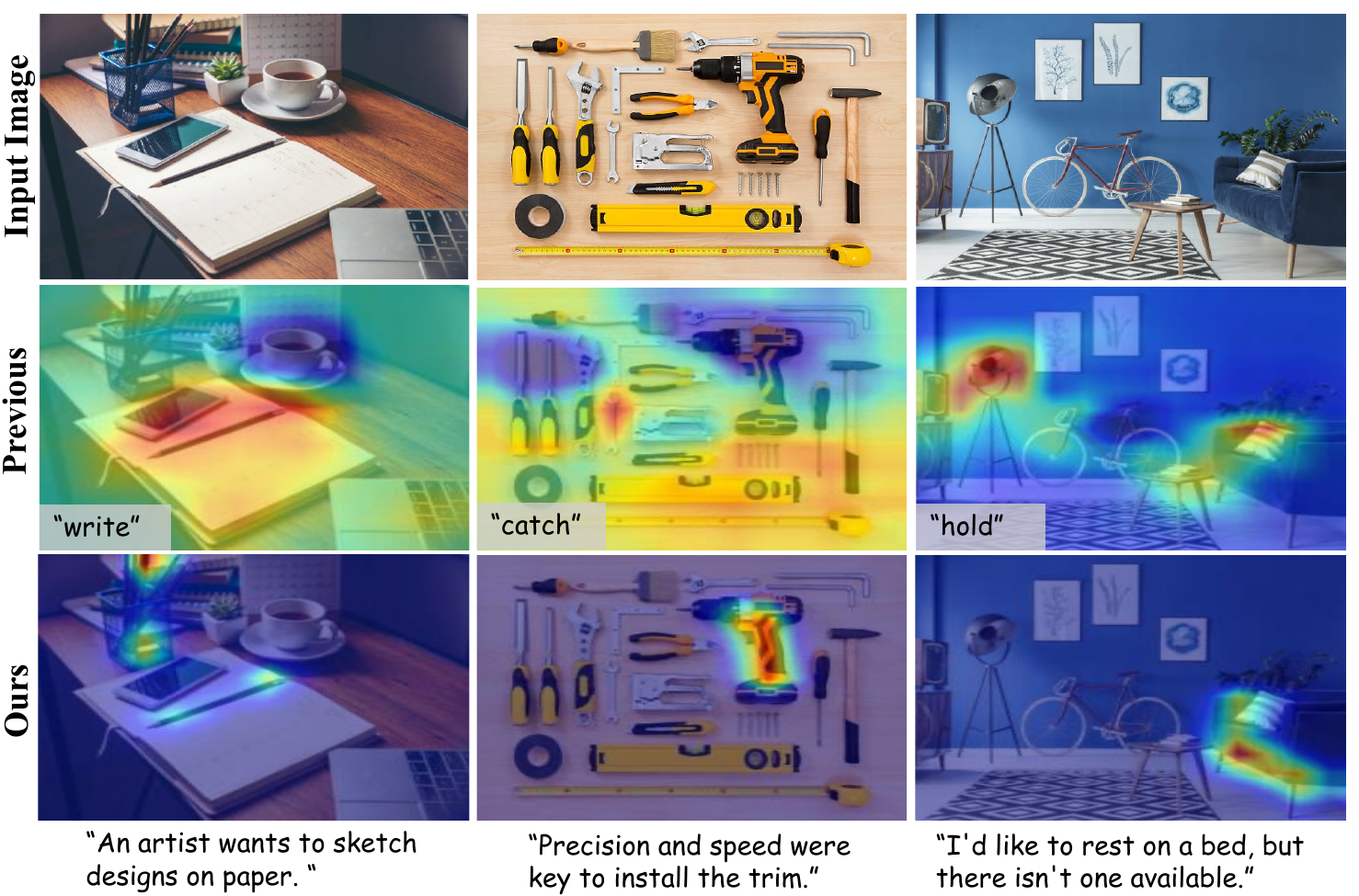}
\caption{Different from previous works only using naïve action labels for affordance grounding, WorldAfford can derive affordance knowledge from LLMs and precisely localize the affordance regions corresponding to natural language instructions. In this way, our framework can work effectively in complex open-world environments. The results in the second row are from Cross-view-AG+~\cite{luo2023grounded}.
}
\vspace{-3mm}
\label{fig:teaser}
\end{figure}

A critical challenge in affordance grounding is instruction comprehension, which means that the embodied agent should understand the human instructions and reason about the actions it is going to perform, which emphasizs active interaction between humans and their environment rather than passive detection.
Furthermore, the agent should analyze which tools in the usage environment can accomplish the given instructions and localize the interaction regions (\textit{i.e.}, affordance regions) on the objects. 
These challenges are expected to be alleviated through using large-scale vision-language foundation models. 
Unfortunately, the currently available models~\cite{li2022blip, zhu2023minigpt, zhang2023multimodal, liu2023llava, cong2023learning, radford2021learning} have not performed satisfactorily on this particular task.

Most recent works~\cite{luo2022learning, zhai2022one, li2023locate, luo2023grounded} attempt to transfer knowledge from exocentric images of an object in an active state to egocentric images where the object is not being used.
They have achieved impressive progress, making dataset collection easier and learning that the affordance region of an object changes dynamically depending on the different given instructions. 
Nevertheless, current approaches can only support simple action labels (\textit{e.g.,} ``\texttt{catch}'' shown in~\cref{fig:teaser}) as input instructions, which cannot express complex human goals.
Besides, these methods can only identify the affordance region of a single object in object-centric images, overlook object context, and still fall short in localizing the affordance regions of multiple objects in complex scene images for practical applications in the real world.
In this paper, for the first time, we introduce a new task of affordance grounding based on natural language instructions, extending affordance grounding from previously using simple action labels to complex natural language instructions.
This new task moves toward real-world applications with significant implications on Embodied AI. 
For this task, we propose a novel framework, WorldAfford, which integrates the large language model (LLM), Segment Anything model (SAM)~\cite{kirillov2023segment}, and CLIP~\cite{radford2021learning}. 
We first use the LLM to process the natural language instruction. 
To reason about affordance knowledge from the LLM more precisely and logically, we design a novel Affordance Reasoning Chain-of-Thought Prompting (ARCoT) including Object-Oriented Reasoning Prompting and Action-Oriented Reasoning Prompting.
Subsequently, we employ SAM and CLIP to segment and select the objects associated with the actions inferred by the LLM. 
Moreover, a Weighted Context Broadcasting module (WCB) is proposed and integrated into the affordance region localization module.
It allows our framework to focus on more informative objects and to identify affordance regions of multiple objects.
% we optimize LOCATE~\cite{li2023locate} to allow it to identify the affordance regions of multiple objects. 
% \ccm{We use LOCATE~\cite{li2023locate}as our affordance region localization module, and optimize it to focus on more informative objects, and allow it to identify affordance regions of multiple objects.}
To benchmark the new task and validate our framework, we constructed a new dataset, LLMaFF, containing 550 test images with natural language instructions and manually labeled affordance maps.
Experimental results demonstrate that our framework outperforms the previous methods both on the existing AGD20K~\cite{luo2022learning} dataset and the new LLMaFF dataset.
Our main contributions can be summarized as follows:
\begin{itemize}
 \item We introduce a new task of affordance grounding based on natural language instructions, extending affordance grounding from using simple action labels to complex natural language instructions.
 \item We propose a framework for this new task named WorldAfford, which integrates the LLM and other vision models. 
 To reason about affordance knowledge from LLMs, we introduce an Affordance Reasoning Chain-of-Thought Prompting.
In addition, we propose a Weighted Context Broadcasting module, allowing WorldAfford to localize affordance regions of multiple objects.
 \item A new dataset LLMaFF is constructed to benchmark the new task.
 \item We conduct extensive experiments to validate that our model performs state-of-the-art on both the AGD20K dataset and our new LLMaFF dataset. 
\end{itemize}

\section{Related Work}

\noindent \textbf{Affordance Grounding.}
Visual affordance grounding has been intensively explored in the fields of robotics and computer vision~\cite{mi2019object, mi2020intention, zhao2020object, luo2021one, hassanin2021visual, deng20213d, luo2023learning, zhou2023novel, li2023beyond, zhou2022robotic, wang2024multi, wang2022transformer}.
Traditional approaches~\cite{do2018affordancenet, fang2018demo2vec, nguyen2017object, chuang2018learning, lu2022phrase} mainly learn the affordance through fully supervised learning.
Nagaragan \textit{et al.}~\cite{nagarajan2019grounded} learn affordance knowledge by watching human-object interaction videos. 
Luo \textit{et al.}~\cite{luo2022learning} propose a Cross-view-AG knowledge transfer framework for affordance grounding, in which the affordance knowledge is acquired from exocentric human-object interactions, and transfer to egocentric images.
Li \textit{et al.}~\cite{li2023locate} extract object-related information from exocentric images and match it to the objects to localize the affordance regions.
However, such methods use only naive action labels for affordance grounding, which may not meet the requirements of practical applications. In this work, we use flexible natural language as supervision to guide agents in localizing affordance regions of multiple objects in complex scenes images.

\vspace{2mm}
\noindent \textbf{Large language models and Chain-of-Thought.}
% What are LLMs
% Famous LLMs
% Their applications in similar works
% In our work LLM's function
Large language models (LLMs), with their extensive world knowledge, play a central role in enabling embodied agents to interpret and execute tasks from human natural language instructions.
While some studies~\cite{huang2023voxposer, huang2022language, chung2022scaling, ahn2022can}have primarily used LLMs to guide object grasping with robotic arms, focusing on basic object perception without considering the fine-grained shapes, functions, or uses of the objects. Our work differs by using LLMs and the affordance reasoning Chain-of-Thought (ARCoT) method to interpret open-world human instructions and reason about a wide range of objects in the environment. Recent studies~\cite{li2024guiding, wei2022chain, zhang2023multimodal, feng2024towards} find that CoT can dramatically improve the performance of LLMs, particularly when dealing with complex tasks involving reasoning, demonstrate that CoT significantly improves the performance of LLMs in complex reasoning tasks. To the best of our knowledge, we are the first to explore CoT for affordance grounding.

\vspace{2mm}
\noindent \textbf{Vision Foundation Model for Affordance Grouding}
% What is segmentation
% SAM
% Why and how we use SAM.
Vision language models have shown promising results in robotics applications~\cite{huang2023instruct2act, khan2024segment}. Some works~\cite{nguyen2023open, van2023open} use vision-language models for affordance detection in 3D point clouds. They focus on the different affordances of individual objects, do not include human instructions, cannot generalize to unseen objects, and require extensive manual annotation. 
Li \textit{et al.}~\cite{li2023one} propose a vision-language framework to address the problem of one-shot affordance learning. 
Ren \textit{et al.}~\cite{ren2024grounded} combines GroundingDINO~\cite{liu2023grounding} and SAM to segment objects in image based on object names.
Luddecke \textit{et al.}~\cite{luddecke2022image} design CLIPSeg using the CLIP as a backbone, expanded with a Transformer-based decoder, to enable segmentation based on image or text inputs.
In contrast to the above work, we use CLIP for semantic understanding and SAM for spatial understanding to select objects related to language instruction. 
This combination allows for the accurate identification of objects as dictated by textual input.

\vspace{2mm}
\noindent \textbf{Affordance Grounding Dataset}
Affordance grounding~\cite{fang2018demo2vec,chen2023affordance, luo2023learning, luo2022learning, li2023locate, luo2023grounded,nagarajan2019grounded} has traditionally focused on datasets such as AGD20K~\cite{luo2022learning} and OPRA~\cite{fang2018demo2vec}
mainly for single actionable object scenarios. 
Recently, Hadjivelichkov \textit{et al.}~\cite{hadjivelichkov2023one} introduce the UMD-i dataset, which focuses on single objects and uses pixel-level labels for training, mainly for one-shot affordance learning. Nguyen \textit {et al.}~\cite{inproceedings} propose the IIT-AFF dataset, which assigns affordance labels to each pixel in its images, lacks semantic information about affordance and uses only image inputs for supervised learning. To address these limitations, we present the LLMaFF dataset, which contains scenes with multiple objects and complex language instructions.

% problem definition
% llm and ARCoT, explain why. do not the failure of GPT3.5 for generating results.
% llms and vlms
% the network
% the dataset, alternative.

\section{Task Definition and LLMaFF Dataset}
% Given a human instruction and an image I, we aim to search the affordance of the objects in the image most preferred to afford the human instruction. 
% The affordance offered by the same image are not fixed in quantity, location, and specific area, which may change with different human instruction. 
% %Different from the previous works focusing on explicit action labels, open world affordance grounding is more general, flexible, and useful in practice.
% In comprison, previous methods~\cite{luo2023grounded, li2023locate, luo2022learning, hadjivelichkov2023one} can't understand human instruction, and groud affordance from fixed single objects, and when the image has multiple objects, many unrelevant affordance regions may be highlighted, which may guide the agent to interact with objects unrelated to human instructions.

Given an image $\bm{I}$ and a natural language instruction $t$, affordance grounding based on natural language instructions aims to localize the interaction regions of objects in the scene image and the instruction can be completed through these interactions.
Compared to the setting in previous works~\cite{luo2023grounded, li2023locate, luo2022learning, hadjivelichkov2023one}, affordance grounding based on natural language instructions is more oriented towards practical applications in the real world since there is no restriction on the number of objects in the image and complexity of the input instructions.

\begin{figure}[h]
\centering
\includegraphics[width=1\linewidth]{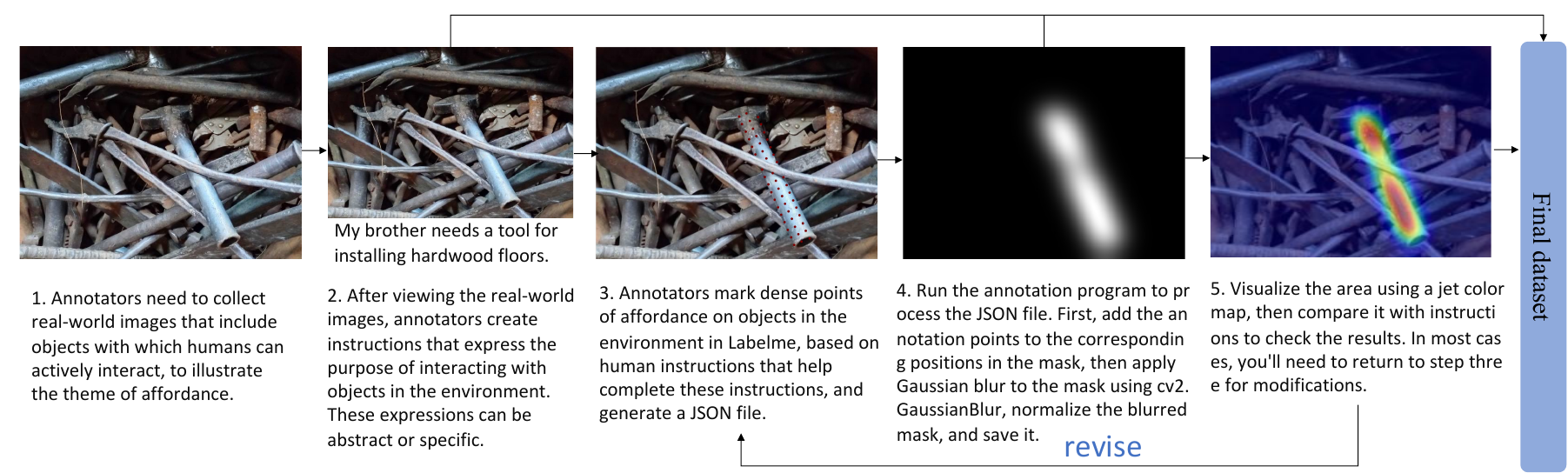}
\caption{Data Collection Pipeline for our WorldAfford benchmark.}
\label{fig:dataset_collection}
\end{figure}

\begin{figure}[h]
\centering
\includegraphics[width=1\linewidth]{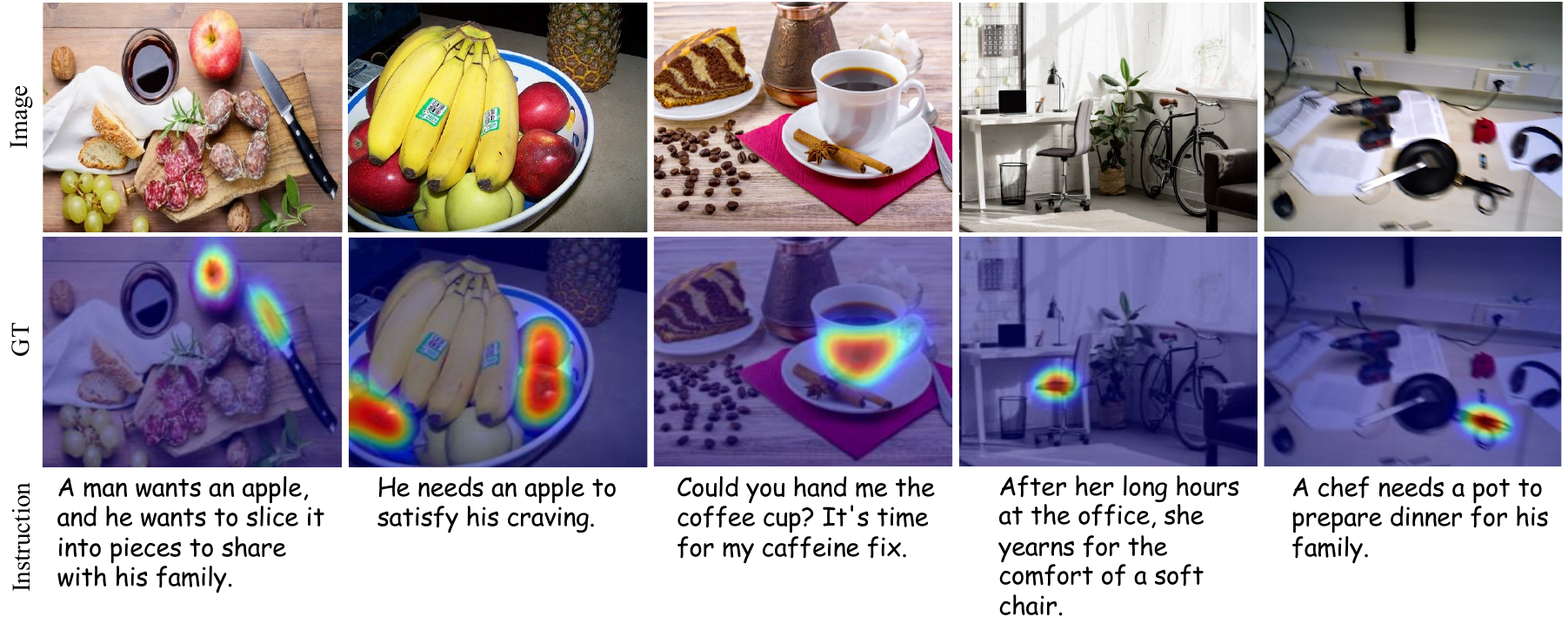}
\caption{Examples from the LLMaFF dataset. The first row presents the images from diverse environments. The second row shows the ground-truth affordance maps, and the third row shows the corresponding natural language instructions.}
\label{fig:dataset}
\end{figure}

To facilitate and benchmark this new task, we construct a new dataset, LLMaFF, consisting of 550 complex environmental images with natural language instructions and manually labelled affordance maps.
The data collection pipeline is shown in~\cref{fig:dataset_collection}.
Annotators need to collect real-world images, create instructions that express the purpose of interacting with objects in the environment, and annotate ground truth (GT) based on these instructions.
The source images of our dataset are primarily sampled from IIT-AFF~\cite{inproceedings}.
Due to the limited object categories of IIT-AFF, we augment the dataset with the images sampled from Ego4D~\cite{grauman2022ego4d} and the Internet. 
\begin{figure}[tb]
\centering
\includegraphics[width=1\linewidth]{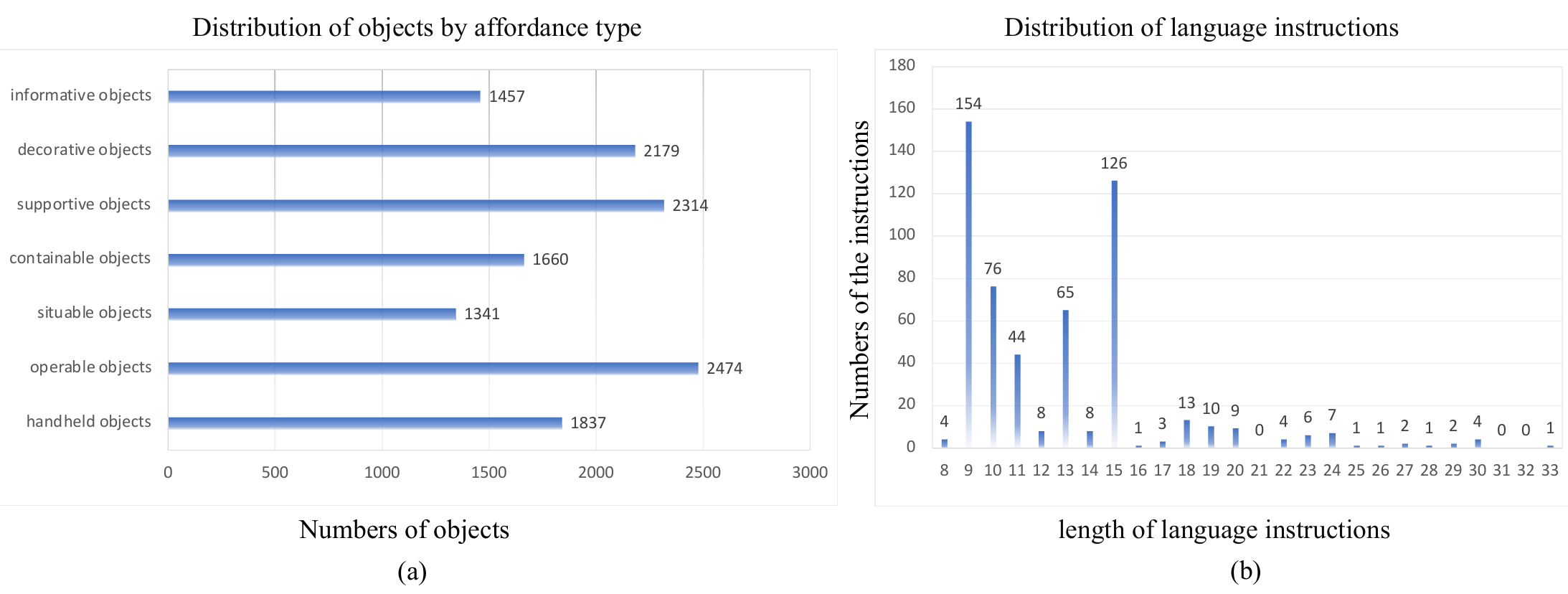}
\caption{Statistics of objects and instructions in WorldAfford. (a) Distribution of objects by affordance type. (b) Distribution of language instructions}
\label{fig:dataset_statistics}
\end{figure}
AGD20K~\cite{luo2022learning} annotates the affordance regions with sparse points and applies a Gaussian kernel to generate ground truth. In contrast, we employ dense points to annotate the affordance map of multiple objects based on the language instructions, which requires careful identification of the objects and their interactions. 
We find that the density and distribution of the points have a significant impact on the labelling results, thus ensuring a uniform distribution of annotation points across multiple objects is crucial to avoid certain regions in the affordance map appearing blank or with faint heat.
Based on the affordances of objects in the environment, we categorized them into eight types: handheld objects(1837), operable objects(2474), situable objects(1341), containable objects(1660), supportive objects(2314), decorative objects(2179), and informative objects(1457). We also conducted a statistical analysis of the length of human language instructions in the dataset, as shown in~\cref{fig:dataset_statistics}.
We demonstrate some examples from the LLMaFF dataset in~\cref{fig:dataset}.
% The framework of our proposed Dreamafford is illustrated in Figure 2. First, we introduce the problem formulation in section 3.1, Second, in section 3.2, we extract affordance knowledge from LLMs via affordance reasoning and chain-of-thought prompting. Next, vision foundation models alongside SAM are employed to detect and segment environmentally situated objects that correspond with the affordance insights derived from LLMs. In parallel, the Human Skill Transfer Network is utilized to implement methods of using these objects, as inferred from LLM outputs, and to delineate affordance regions with heat maps, visualizing the affordance regions through heat maps. Lastly, we detail the dataset used to train and test this Skill Transfer Network. 
% This approach underscores our framework's innovative integration of visual perception and language-based reasoning to facilitate advanced affordance grouding capabilities. 
\section{WorldAfford Framework}
\begin{figure}[tb]
\centering
\includegraphics[width=1\linewidth]{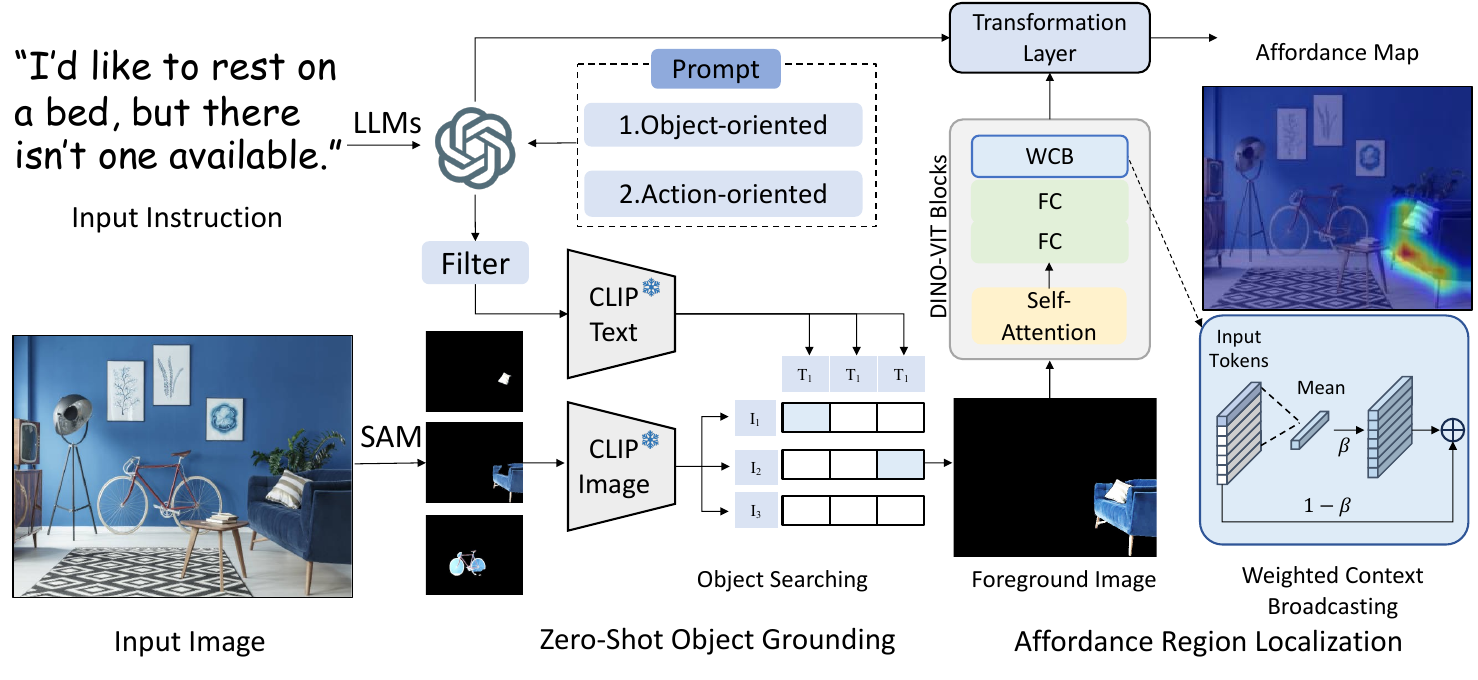}
\caption{Framework of WorldAfford. We first derive affordance knowledge from LLMs with the proposed affordance reasoning chain-of-thought. Subsequently, we utilize SAM and CLIP for zero-shot object grounding. Moreover, we design a WCB module and integrate it into the affordance region localization module to localize the affordance regions of multiple objects.} 
\label{fig:framework}
\end{figure}

We propose WorldAfford as a general framework for affordance grounding incorporating complex instruction understanding and multi-object affordance localization at a very low training cost.
We first use the LLM~\cite{achiam2023gpt} to analyze the given instruction and derive affordance knowledge through the affordance reasoning chain-of-thought prompting. 
Subsequently, we utilize SAM~\cite{kirillov2023segment} and CLIP~\cite{radford2021learning} to implement zero-shot multi-object grounding, segmenting and selecting the objects associated with the sub-actions provided by the LLM.
Furthermore, we design a Weighted Context Broadcasting (WCB) and integrate it into the affordance region localization module to localize the affordance regions of multiple objects.
The overall framework of WorldAfford is illustrated in~\cref{fig:framework}.

%\subsection{Affordance Knowledge from LLMs} 
\subsection{Affordance Reasoning Chain-of-Thought Prompting}
\label{sec:affordance_cot}

% \begin{wrapfigure}{l}{0.5\textwidth}
% \includegraphics[width=0.5\textwidth]{figures/cot_new.pdf}
% \vspace{-3mm}
% \end{wrapfigure}
When humans receive an instruction, they initially consider which tools (objects) might facilitate the task and decompose the complex instruction into a series of simpler actions for execution.
In this paper, we prompt the LLM to mimic this aspect of human cognitive processing, and reason about relevant affordance information using the extensive world knowledge learned from large-scale data. 

Rather than relying on direct inference, our approach employs a straightforward and effective chain-of-thought prompting to enhance the capabilities of LLMs in affordance reasoning. 
The proposed chain-of-thought prompting consists of two primary strategies: (1) \textbf{object-oriented reasoning prompting}, and (2) \textbf{action-oriented reasoning prompting}. 
It enables the LLM to derive crucial object and action information from natural language instructions. 
Furthermore, our method allows the LLM to identify alternative tools when the optimal one is not available, demonstrating its adaptability to various scenarios.
\subsubsection{Object-Oriented Reasoning Prompting.} 
\label{object_prompting}
% We first utilize the LLM to reason about the possible objects that can afford the given instruction. 
% \noindent\textit{\scriptsize{
% Output: Chair..., Hammock..., Blanket and Pillows...}}
% \vspace{1mm}

% %Considering the possible co-occurrence of multiple actions in the given instruction and the diversity of the objects that can afford each action, the LLM is requested to output a set of object categories with the size $k$:
% Considering that multiple objects are likely to be necessary for completing the instruction, the LLM is requested to output a set of object categories $\mathcal{O}$:
% \begin{equation}
%     \mathcal{O} = \text{LLM}(k, t, p_{obj}),
% \end{equation}
% where $k$ indicates the size of the object set and $t$ denotes the given natural language instruction.

% The overview of affordance reasoning chain-of-thought prompting is illustrated in~\cref{fig:arcot}.

% The prompt $p_{obj}$ for object-oriented reasoning is specifically designed as follows:

% \vspace{1mm}
% \noindent\textit{\scriptsize{
% Prompt: What are the [\#$k$] most common objects that can be used if [\#$t$]?}}

% The object-oriented reasoning prompts the LLM to provide diverse objects suitable for an action. 
% Moreover, it associates alternative tools in case the best tool does not exist in the environment, which facilitates the accomplishment of the instruction.
% These inferred object categories from the LLM are further utilized for subsequent action-oriented reasoning.
We first utilize the LLM to reason about the possible objects that can afford the given instruction.  
Considering that multiple objects are likely to be necessary for completing the instruction, the LLM is requested to output a set of object categories $\mathcal{O}$:
\begin{equation}
\label{eq:eq1}
    \mathcal{O} = LLM(k, t, p_{obj}),
\end{equation}
where $k$ indicates the size of the object set and $t$ denotes the given natural language instruction.
The prompt $p_{obj}$ for object-oriented reasoning is specifically designed as follows:

\vspace{1mm}
\noindent\textit{\footnotesize{
Prompt: What are the [\#$k$] most common objects that can be used if [\#$t$]?}}

\noindent\textit{\footnotesize{
Output: Chair..., Hammock..., Blanket and Pillows...}}

The object-oriented reasoning prompts the LLM to provide diverse objects suitable for an action. 
Moreover, it associates alternative tools in case the best tool does not exist in the environment, which facilitates the accomplishment of the instruction.
These inferred object categories from the LLM are further utilized for subsequent action-oriented reasoning.

We designed a filter function based on the large model's output about object descriptions to filter out excessive explanatory text, which sometimes includes irrelevant objects, hindering the subsequent object search.

\begin{wrapfigure}{r}{0.45\textwidth} % 降低宽度比例
\vspace{-5mm} % 减小负空间
\centering
\includegraphics[width=0.45\textwidth]{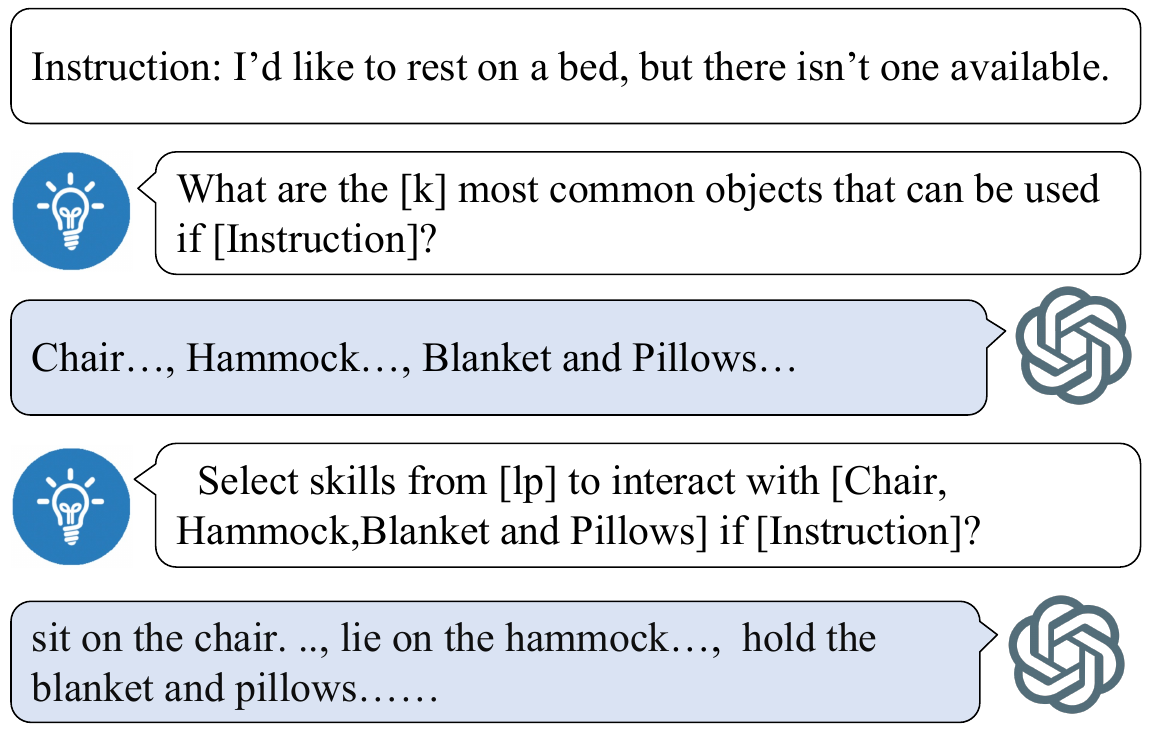}
\caption{The affordance reasoning Chain-of-Thought prompting.}
\vspace{-8mm} % 减小负空间
\label{fig:arcot}
\end{wrapfigure}
\subsubsection{Action-Oriented Reasoning Prompting.}
Different from previous work on locating the affordance for single action labels, to address the complex natural language instructions, we utilize the powerful prior knowledge of the LLM to decompose a complex instruction into several simple sub-actions.
Given a pre-defined predicate list,  we prompt the LLM to select the appropriate predicates from it and assign these predicates to the objects in the object set $\mathcal{O}$.
A set of sub-actions $\mathcal{A}$ is generated as follows:
\begin{equation}
    \mathcal{A} = \text{LLM}(\mathcal{O} , l_p, t, p_{act}),
\end{equation}
where $l_p$ indicates the pre-defined predicate list and $t$ denotes the given instruction.
Each sub-action consists of a predicate and an object.
The prompt $p_{act}$ for action-oriented reasoning is specifically designed as follows:

\vspace{1mm}
\noindent\textit{\footnotesize{
Prompt: Select skills from [{\#$l_p$}] to interact with the above [\#$\mathcal{O}$] to help me if [\#$t$]?}}

\noindent\textit{\footnotesize{
Output: sit on the chair..., lie on the hammock..., hold the blanket and pillows...}}
\vspace{1mm}

The inferred sub-actions from the LLM are further utilized as the input of the following affordance region localization module.

% \noindent\textbf{Environment adaption Knowledge Prompting.} To empower agents with the ability to adapt flexibly to the current environment and creatively use existing items or resources to solve problems or complete tasks. It involves finding alternatives in the absence of traditional tools, thus requiring significant flexibility, creativity, and adaptability. We have designed the following text prompt: 

% Ultimately, we obtain the affordance knowledge inferred by LLMS, and we perform knowledge aggregation, extracting object-level knowledge, filtering out a few knowledge unsuitable for human instructin, and transfer the high-level human instrction to a closen set of actionalable skills to guide embodied agents in learning, which is also called extracting actionable knowledge for embodied agents~\cite{huang2022language}, we prompt LLMS to generate reasonable action plans for embodied agents based on human instructions, followed by the implementation of visual cues for the actions of the embodied agents.
We use the LLM to extract object-level knowledge and aggregate action-level knowledge. 
In the inference process of the LLM, irrelevant information in the natural language instruction is ignored and the highly abstract instruction is transformed into a series of executable sub-actions.
The powerful reasoning ability and adaptive results of the affordance reasoning chain-of-thought facilitate the subsequent zero-shot object grounding and the affordance region localization. 

\subsection{Zero-shot Multiple Object Grounding}
To integrate the affordance knowledge provided by the LLM with visual information about the environment, our approach leverages the capabilities of Segment Anything Model (SAM)~\cite{kirillov2023segment} and CLIP~\cite{radford2021learning} to effectively ground the relevant objects in the scene image according to the given natural language instruction. 
The impressive zero-shot performance of SAM and CLIP enables our framework to precisely localize objects across the open world without the need for extensive and expensive training on large-scale datasets.

Initially, SAM produces $N$ segmentation masks for the input image.
These masks, while precisely segmented, lack semantic labels and unavoidably contain irrelevant objects.
In order to obtain the object masks that are relevant to the given instruction, CLIP is integrated to compute the similarity between the visual appearance of the masks and the object categories provided by the LLM. 
We extract the corresponding regions from the original image $\bm{I}$ based on the segmentation masks. 
Subsequently, the cropped regions $\bm{m}$ are encoded by the CLIP image encoder $\textrm{E}_{image}$ while the textual object categories $\bm{o}$ are encoded by the CLIP text encoder $\textrm{E}_{text}$.
The probability $p$ of the mask being classified as the $i$-th object category can be formulated as:
\vspace{-1mm}
\begin{equation}
\centering
p=\frac{\exp( \textrm{sim}(\textrm{E}_{image}(\bm{m}), \textrm{E}_{text}(\bm{o}_i))/\alpha)}{\sum_{o_i\in \mathcal{O}} \exp( \textrm{sim}(\textrm{E}_{image}(\bm{m}), \textrm{E}_{text}(\bm{o}_i))/\alpha)},
\label{eq:loss}
\end{equation}
where $\textrm{sim}(,)$ denotes the cosine similarity function and $\mathcal{O}$ indicates the set of object categories from the LLM. 
The scaling factor $\alpha$ is set to 0.1 in practice.
We establish a boundary to determine whether the masks from SAM are valid.
The masks with probability $p$ above the boundary are identified as valid masks. 
With these active masks, we construct a full-view segmentation mask in which the region covered by the valid masks is viewed as foreground, while the remaining area is considered as background. 
This full-view mask is the same size as the input image and is further used for affordance region localization.

\subsection{Affordance Region Localization}
\label{sec:method:actionnetwork}

To localize the affordance region of the objects in the image corresponding to the given instruction, we employ LOCATE~\cite{li2023locate} and enhance the grounding performance through two crucial improvements:
(1) We use the full-view mask resulting from zero-shot multi-object grounding to preserve the foreground and mask off the background as the input, rather than the entire image.
(2) We propose a weighted context broadcasting (WCB) module, seamlessly integrating it into DINO-ViT~\cite{caron2021emerging} to enable the model to prioritize informative objects. 
With these improvements, our approach outperforms the original LOCATE and can localize multiple affordance regions with the knowledge provided by the LLM.

We utilize the full-view mask from zero-shot multi-object grounding to mask off the irrelevant objects in the image. 
The relevant objects are preserved and the image is forwarded into DINO-ViT to extract deep part-aware features. 
% Different from LOCATE, we finetune the DINO-ViT backbone during training.
Different from LOCATE, we design a Weighted Context Broadcasting (WCB) module inspired by Context Broadcasting~\cite{hyeon2023scratching} and incorporate it into DINO-ViT as demonstrated in \cref{fig:framework}.
Given a sequence of $N$ patch tokens, the WCB module combines the average context tokens with the
input tokens in a weighted manner as follows:
\begin{equation}
    \text{WCB}(x_i) = x_i * \beta + \frac{1}{N}\sum_{j=1}^{N} x_j*(1-\beta),
\end{equation}
where the weight $\beta$ is an empirically determined hyperparameter. 
In order to improve the model's capability to perceive multiple objects, it is expected that the attention maps in the self-attention modules of DINO-ViT are dense rather than sparse.
It has been discussed in~\cite{hyeon2023scratching} that aggregating the average context token can facilitate the self-attention modules to learn dense attention maps.
However, such simple aggregation makes training difficult since the target attention is unknown and uncertain.
To solve this issue, we introduce a weight to balance the aggregation. 
With the proposed WCB, the target attention is easier to learn and the model can focus on more informative objects. 
The experiment in~\cref{sec:quantitative_results} also demonstrates that our approach outperforms the previous works \cite{luo2023grounded} in terms of affordance grounding for objects.

The feature maps generated by DINO-ViT are further refined by a transformation layer including a feed-forward layer and two subsequent convolutional layers. 
We follow the training strategy of LOCATE~\cite{li2023locate} to transfer affordance knowledge from exocentric images to egocentric images.
To predict the affordance maps, a convolutional layer with a window size of $1\times1$ is utilized to project the channel number to the total number of the action categories in the pre-defined predicate list $l_p$.
We aggregate the affordance maps corresponding to the action categories provided by the LLM and normalized them to limit the activation values in the map between 0 and 1 as the final output.

\section{Experiments}
\label{sec:blind}
% 提出的网络（基于locate那篇文章修改的）是用于单个物体的affordance识别，并且在AGD20K上训练，与之前的进行结果和参数对比。
% 针对多物体以及人类指令输入，没有相关的测试集，所以基于互联网的生活照片以及coco，ego4d截取照片建立一个100或者更多的测试集，标注方法以及metrics和之前的一样，并给出数值结果以及消融后的结果。
% 如有时间，进行大模型推理能力探索，提出更抽象的问题，目前发现它无法处理比如说想要创造一些新的东西这种指令。
\subsection{Datasets and Evaluation Metrics}
We conduct experiments on the AGD20K~\cite{luo2022learning} dataset and our proposed LLMaFF dataset. 
AGD20K dataset stands out as the only large-scale dataset containing 20,061 demonstration images and 6,060 object images for training. 
We select the test set of 1,675 images to assess the performance of our method in the task of affordance grounding guided by a single action label.
LLMaFF dataset includes 550 complex environment images with natural language instructions and affordance maps. 
We conduct experiments on the LLMaFF dataset to evaluate the performance of our method in the task of affordance grounding based on natural language instructions.
We use KLD~\cite{bylinskii2018different}, SIM~\cite{swain1991color}, and NSS~\cite{peters2005components} as metrics to measure the correspondence between the predicted affordance map and the ground truth. Specifically, KLD quantifies the divergence between the predicted affordance maps and the distribution of ground truth images, SIM assesses the similarity between the affordance maps and the ground truth, and NSS measures the agreement between the prediction and the ground truth.
\subsubsection{Training strategy and comparison fairness}
Only the affordance region localization module requires training, while other modules are frozen. 
The training is only performed on AGD20K with the same settings as the baselines at a low trainging cost.
Overall, we keep the comparison as fair as possible.
\begin{table}[t!]
\centering
\small
\caption{Comparison of WorldAfford with other state-of-the-art affordance grounding methods on AGD20K. The best numbers are highlighted in \textbf{bold}.}
\vspace{0mm}
\resizebox{\textwidth}{!}{
\begin{tabular}{>{\centering\arraybackslash}p{3.5cm}>{\centering\arraybackslash}p{3.5cm}>{\centering\arraybackslash}p{2cm}>{\centering\arraybackslash}p{2cm}>{\centering\arraybackslash}p{2cm}}
   % \begin{tabular}{ccccc}
\toprule
Approach & Input Instruction & KLD$\downarrow$ & SIM$\uparrow$ & NSS$\uparrow$ \\
\midrule
Hotspots \cite{nagarajan2019grounded}& Action Label & 1.773 & 0.278 & 0.615 \\
Cross-view-AG\cite{luo2022learning}&Action Label & 1.538 & 0.334 & 0.927 \\

Affcorr \cite{hadjivelichkov2023one}&Action Label & 1.407 & 0.359 & 1.026 \\
LOCATE \cite{li2023locate}&Action Label & 1.226 & 0.401 & 1.177 \\
Cross-view-AG+ \cite{luo2023grounded}& Action Label & 1.213 & 0.403 & 1.242 \\
WorldAfford(ours)&Action Label&\textbf{1.201}&\textbf{0.406}&\textbf{1.255}\\
\bottomrule
\end{tabular}
}
\label{tbl:singleactionlabel}
\vspace{-3mm}
\end{table}

\begin{table}[t!]
\centering
\small
\caption{{Comparison on LLMaFF dataset. We manually select labels for the other methods to comparison with them. WorldAfford outperforms all previous methods across all evaluation metrics. The best results are highlighted in \textbf{bold}.}}.
\vspace{0mm}
\resizebox{\textwidth}{!}{%
\begin{tabular}{>{\centering\arraybackslash}p{3.5cm}>{\centering\arraybackslash}p{3.5cm}>{\centering\arraybackslash}p{2cm}>{\centering\arraybackslash}p{2cm}>{\centering\arraybackslash}p{2cm}}
\toprule
Approach & Input Instruction & KLD$\downarrow$ & SIM$\uparrow$ & NSS$\uparrow$ \\
\midrule
Cross-view-AG+\cite{luo2023grounded}&Action Label& 2.927 & 0.123 & -0.194 \\
Cross-view-AG \cite{luo2022learning}&Action Label& 2.887 & 0.119 & 0.118 \\
LOCATE \cite{li2023locate}&Action Label& 1.958 & 0.212& 1.713 \\
WorldAfford(ours)&Natural Language&\textbf{1.163}&\textbf{0.386}&\textbf{2.819}\\
\bottomrule
\end{tabular}
}
\label{tbl:multipleobjects}
\vspace{-3mm}
\end{table}

\subsection{Implementation Details.} We use GPT-4~\cite{achiam2023gpt} as the large language model, while both the CLIP~\cite{radford2021learning} and the Segment Anything Model (SAM)~\cite{kirillov2023segment} implement object matching and segmentation in a zero-shot fashion. The affordance information is extracted from the output of the large language model by removing most of the irrelevant text to allow the CLIP to more accurately localize the position of objects. The affordance region localization module is trained on a RTX 3090 GPU. 
\begin{figure*}[t!]
    % \captionsetup{font=small}
    \centering 
    \includegraphics[width=1.0\textwidth]{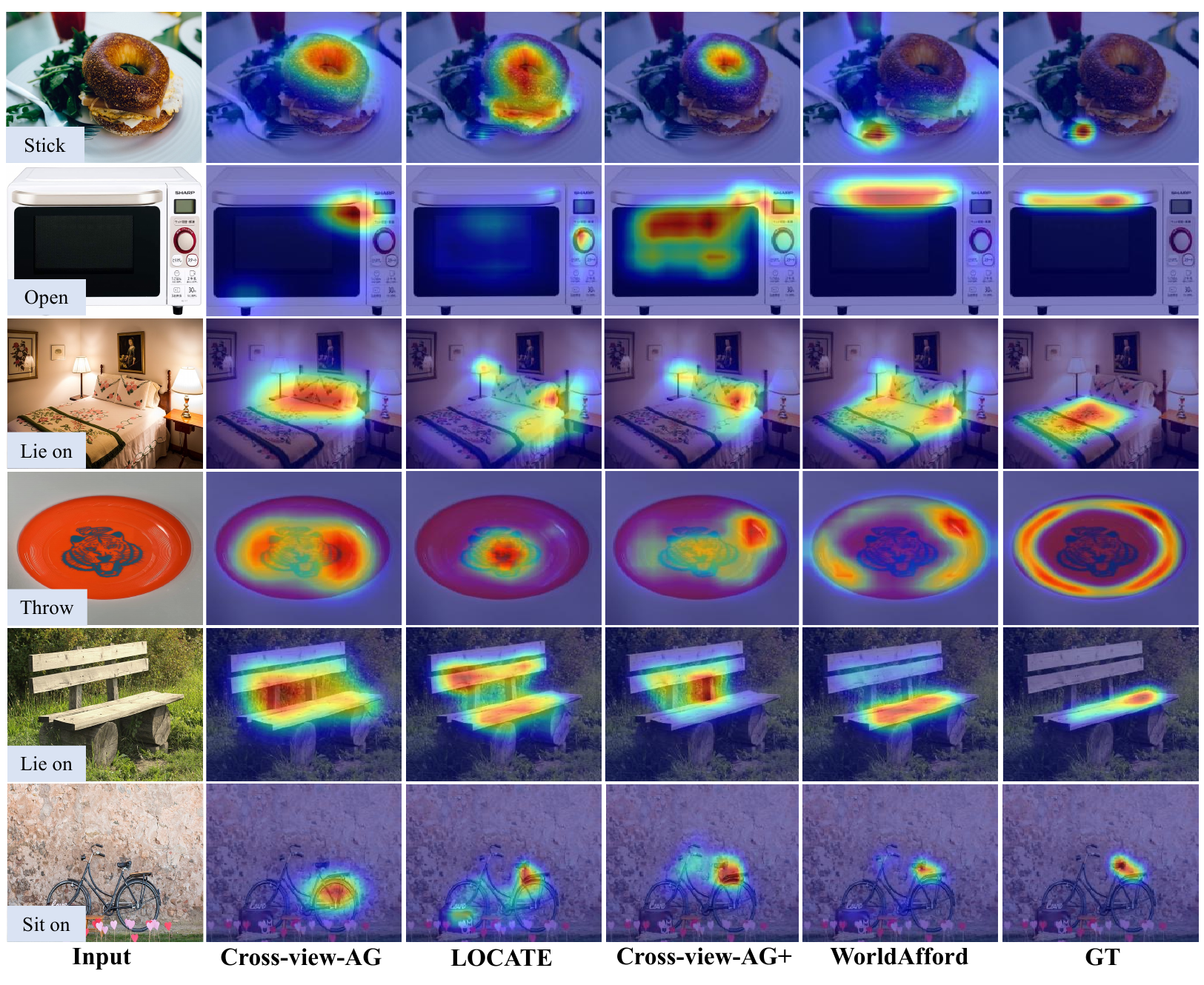}
    \vspace{-7mm}
    \caption{Visual comparison on the AGD20K dataset. Compared to previous methods, our method can infer more precise affordance maps.}
    \label{fig:singleactionlabel}
    \vspace{-5mm}
\end{figure*}
We load the pre-trained DINO-ViT~\cite{caron2021emerging} model and finetune the features it extracts from images.
We set the weight $\beta$ to 0.88, and the number $k$ in~\cref{eq:eq1} is set to 3.
We use a learning rate of 0.005, a decay factor of 5e-4, a batch size of 16, and train the affordance region localization module for 35 epochs.

\subsection{Quantitative results} 
\label{sec:quantitative_results}
For comparison with the previous affordance grounding approaches, we first validate our framework on the AGD20K dataset.
AGD20K is widely used in the affordance grounding approaches~\cite{luo2022learning, li2023locate, luo2023grounded}, which employ a simple action label to localize the affordance regions of a single object in the object-centric images. 
Since WorldAfford uses natural language instructions as the input, a direct comparison is difficult to achieve.
To solve this problem, we only use action labels as input to the affordance region localization module and compare it to these approaches. 
The results in~\cref{tbl:singleactionlabel} show that even in this simplified setting, our approach still outperforms the previous methods. 
WorldAfford establishes a new state-of-the-art performance in affordance grounding. 
It shows that our weighted context broadcasting module facilitate the framework to focus more on object-oriented information, effectively identifying the affordance regions.
\begin{figure*}[t!]
    % \captionsetup{font=small}
    \centering 
    \includegraphics[width=1.0\textwidth]{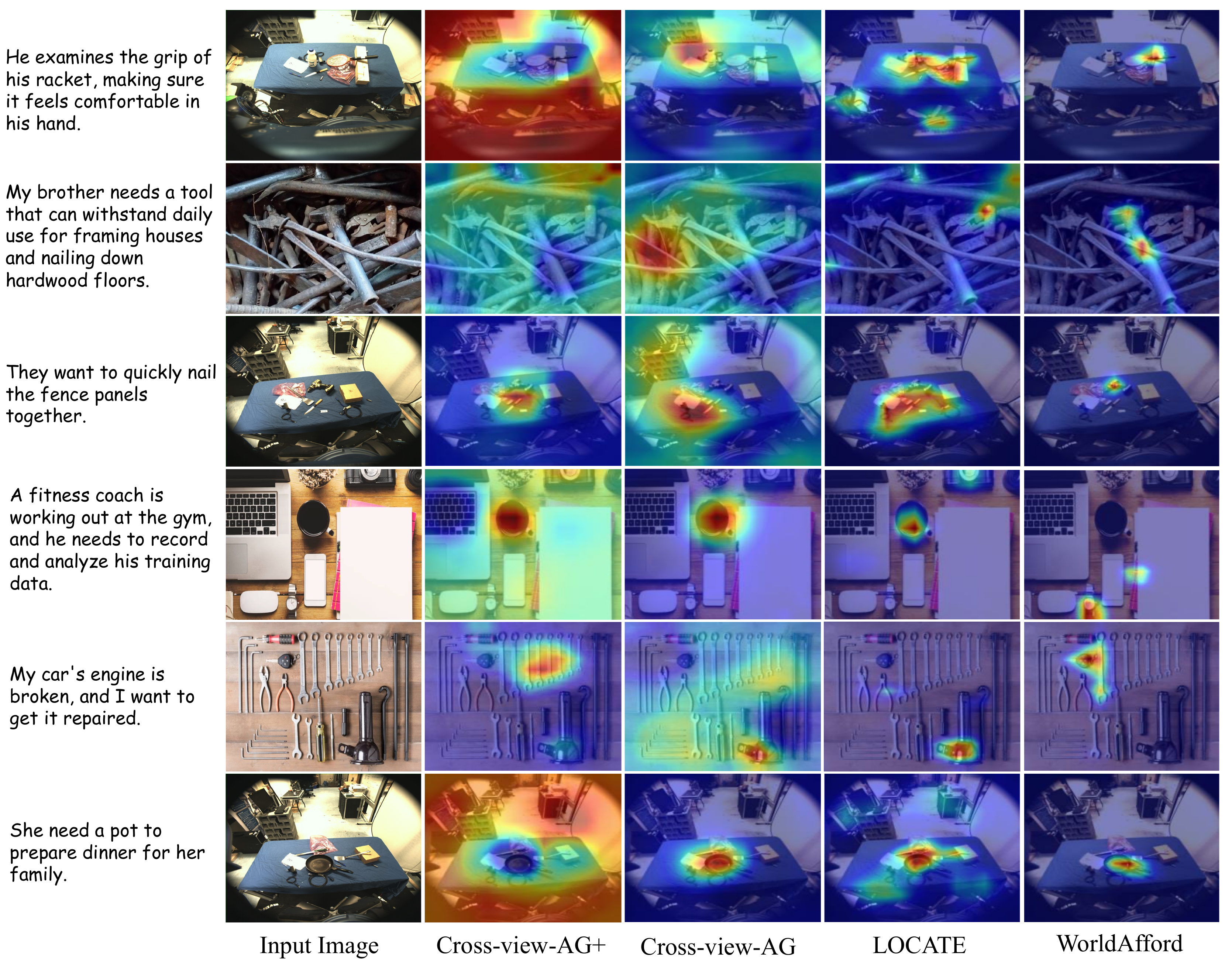}
    % \vspace{-7mm}
    \caption{Visual comparison on the LLMaFF dataset. We manually assign labels to other methods since they cannot adopt the  textual input. The labels, "swing", "carry", "catch", "pick up", "catch", and "carry" correspond to the first through sixth rows respectively. The previous approaches cannot predict the affordance regions based on the textual instructions, whereas our method successfully identifies the affordance regions corresponding to natural language instructions and performs outstandingly in the complex environments.}
    \label{fig:results_on_LLMaFF}
    \vspace{-5mm}
\end{figure*}

We further conduct the experiments on the new LLMaFF dataset to evaluate the performance of our method on the new task. 
Since the previous methods~\cite{luo2022learning, li2023locate, luo2023grounded} cannot directly adopt the textual instruction as input, we manually select the appropriate action labels for comparison. 
The results shown in~\cref{tbl:multipleobjects} demonstrate that our method can effectively localize the affordance regions of the objects in the complex scene images.
While Cross-view-AG+~\cite{luo2023grounded} achieves outstanding results on AGD20K, its performance on the LLMaFF dataset is less impressive. 
This indicates a decrease in its ability to accurately localize the affordance regions, as shown by its negative score (-0.194) on the NSS indicator.
There are two explanations for the difference in the performance of Cross-view-AG+ between the two datasets: 
1) The complexity of the new task, with numerous potential disruptions, presents a major challenge.
2) Cross-view-AG+ likely overfit the AGD20K dataset.
The results in~\cref{tbl:multipleobjects}  show a significant decrease in the performance of Cross-view-AG~\cite{luo2022learning} and LOCATE~\cite{li2023locate}, which also demonstrate that previous methods cannot localize the affordance regions of objects in the complex scene images.

\subsection{Qualitative results}
We show the qualitative comparisons on AGD20K in~\cref{fig:singleactionlabel}. The affordance regions identified by Cross-view-AG tend to be too large, sometimes including irrelevant regions. 
In contrast, LOCATE prefers to predict smaller regions but often fails to capture the full affordance region of objects. 
Cross-view-AG+ is able to identify the regions associated with the action label but not accurately.
In contrast, out framework, WorldAfford, achieves the new state-of-the-art performance in this simple setting, providing a sharper and more accurate results. 
It demonstrates that the weighted context broadcasting module (WCB) allows the framework to focus more on informative objects, thus capturing more knowledge of objects and localizing more accurately.

The results on the LLMaFF dataset are shown in~\cref{fig:results_on_LLMaFF}. 
We find that Cross-view-AG+ fails to identify affordance regions of multiple objects in the images, resulting in disordered color distribution, and thus cannot provide effective visual information to the agent.
Cross-view-AG also shows some failure cases. It can capture information about objects in the image. 
However, we observe that the information tends to be biased towards objects with larger sample sizes in the training dataset. The results demonstrate the lack of comprehensive understanding of the objects in the image. 
LOCATE can capture the affordance of a few objects.
However, it often activates the affordance regions of some irrelevant objects, and may interpret several objects as a single entity. 
In comparison, our method can predict the affordance maps that are more consistent with natural language instructions, more accurate, and is capable of localizing the affordance regions of multiple objects, thus providing richer visual information.
\begin{figure*}[t!]
    \centering 
    \includegraphics[width=1.0\textwidth]{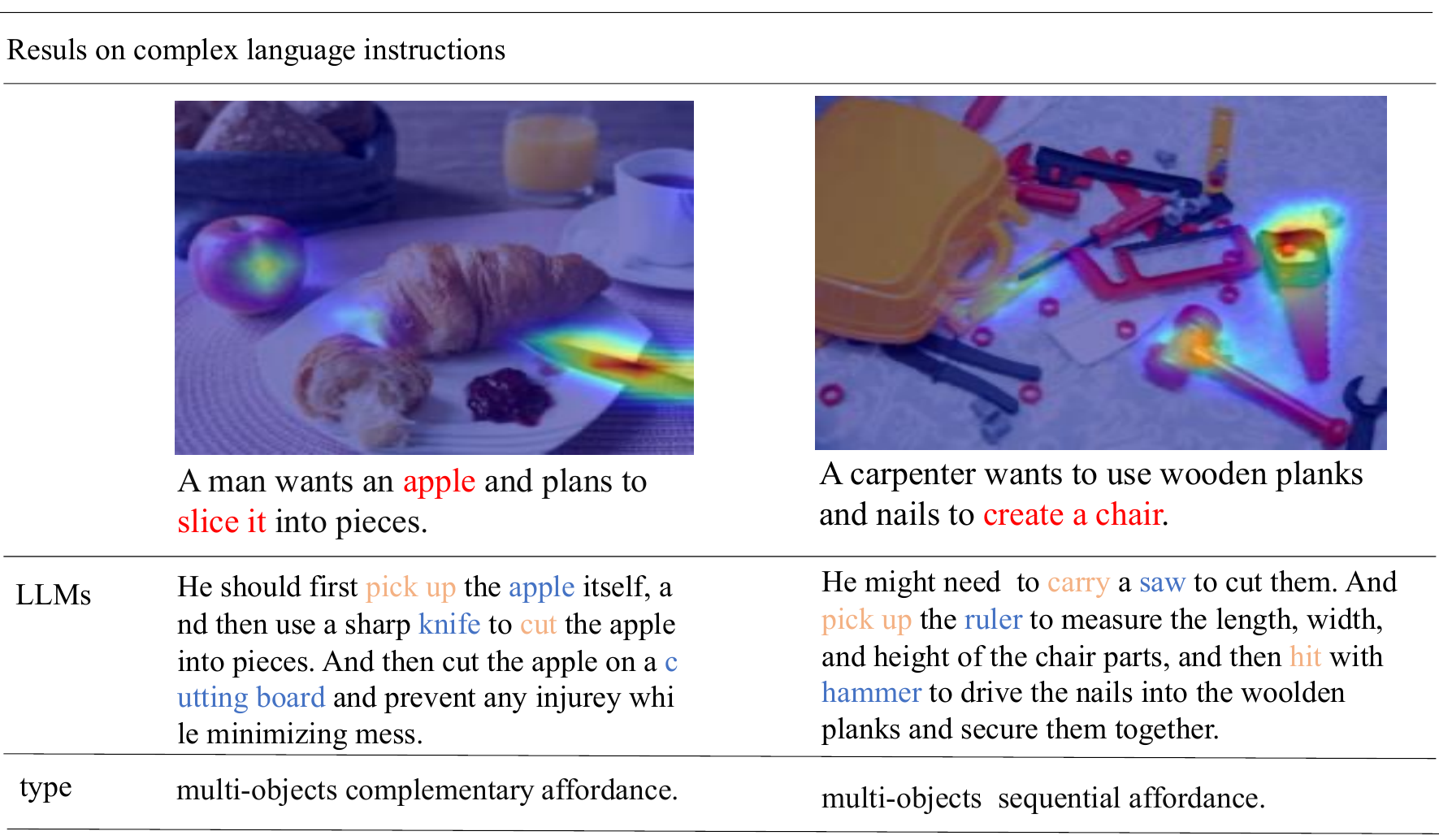}
    % \vspace{-7mm}
    \caption{Affordance results based on difficult language instructions. While previous methods struggle to infer from difficult language instructions, our method demonstrates the capability to comprehend such instructions and accurately identify the affordance regions of multiple objects.}
    \label{fig:two_examples}
    \vspace{-5mm}
\end{figure*}
\subsubsection{Results on complex language instructions}
We explore the challenge of affordance grounding based on complex language instructions that previous methods~\cite{luo2023grounded} cannot understand, as illustrated in~\cref{fig:two_examples}. 
Unlike simple ones, these instructions require a deeper understanding of human knowledge, demonstrating the superior flexibility and creativity of our method. 
WorldAfford successfully identifies intricate affordance regions, and can highlight the complementarity of object interactions, such as using a knife and an apple together for slicing, which provides the embodied agent with detailed visual information that enhances its ability to follow complex instructions involving multiple sub-tasks. 
Our exploration of building a chair from wooden planks and nails illustrates how our method systematically identifies and activates the necessary affordance regions for sawing, measuring, and assembling, providing a comprehensive solution to multi-step tasks. This advance in affordance grounding may open new avenues for robotics and AI applications, significantly enriching the interaction between agents and their environment.
\begin{table}[h]
\centering
\small
\caption{Generalization ability comparison of WorldAfford with other state-of-the-art affordance grounding methods on AGD20K. }
\vspace{0mm}
\resizebox{\textwidth}{!}{
\begin{tabular}{>{\centering\arraybackslash}p{3.5cm}>{\centering\arraybackslash}p{3.5cm}>{\centering\arraybackslash}p{2cm}>{\centering\arraybackslash}p{2cm}>{\centering\arraybackslash}p{2cm}}
   % \begin{tabular}{ccccc}
\toprule
Approach & Input Instruction & KLD$\downarrow$ & SIM$\uparrow$ & NSS$\uparrow$ \\
\midrule
Hotspots \cite{nagarajan2019grounded}& Action Label & 1.994 & 0.237 & 0.577 \\
Cross-view-AG\cite{luo2022learning}&Action Label & 1.787 & 0.285 & 0.829 \\
Affcorr \cite{hadjivelichkov2023one}&Action Label & 1.618 & 0.348 & 1.021 \\
LOCATE \cite{li2023locate}&Action Label & 1.405 & 0.372 & 1.157 \\
WorldAfford(ours)&Action Label&\textbf{1.393}&\textbf{0.38}&\textbf{1.225}\\
\bottomrule
\end{tabular}
}
\label{tbl:generalization}
\vspace{-3mm}
\end{table}
\subsection{Generalization ability and learnable parameters}
To evaluate the generalization ability of our method, we add the results of the unseen test on AGD20K, which is shown in~\cref{tbl:generalization}.
Additionally, all LLMaFF images, including various scenes and many object categories such as nail gun, smartwatch and so on, are unseen in training, which also demonstrates the the superior generalization ability of our method.
We use the the knowledge of foundation models, the training cost is very low, and the comparison of learnable parameters: 120.03M(Cross-view-AG)/82.27M (Cross-view-AG+)/6.5M (LOCATE)/6.5M (WorldAfford).
\subsection{Ablation Study}
\label{sec:ablation_study}
We conduct the ablative experiments on the LLMaFF dataset to validate the effectiveness of the affordance reasoning chain-of thought prompting(ARCoT). The results shown in~\cref{tbl:ablation_study} demonstrate that the object information and the action information derived from the LLM via our affordance reasoning chain-of-thought prompting (ARCoT) can both improve the performance for the task of affordance grounding based on language instructions. 
We also validate that the proposed WCB module can enhance the perception of affordance regions by enabling the model to focus on more informative objects.
Overall, our contributions significantly improve the affordance grounding capabilities of the model and establish a new state-of-the-art performance in the affordance grounding based on natural language instructions task.
To verify our adjustments for masking off irrelevant objects, we conduct experiments on LLMaFF, the results is shown in~\cref{tbl:full_images}.
\begin{table}[h]
\centering
\caption{Ablation results of the proposed modules. LMA denotes the action information associated with the manipulated objects inferred from the LLM. WCB indicates the weighted context broadcasting module. LMO represents the object information inferred from the LLM.}
\vspace{2mm}
\resizebox{\textwidth}{!}{
\begin{tabular}{>{\centering\arraybackslash}p{1.5cm}>{\centering\arraybackslash}p{1.5cm}>{\centering\arraybackslash}p{1.5cm}>{\centering\arraybackslash}p{2cm}>{\centering\arraybackslash}p{2cm}>{\centering\arraybackslash}p{2cm}}
\toprule
LMA & WCB & LMO & KLD$\downarrow$ & SIM$\uparrow$ & NSS$\uparrow$ \\ % 移除了第一个 & 符号
\midrule
 &  &  & 3.073 & 0.105 & -0.059 \\ % 移除了每行的第一个 & 符号
 & \cmark &  & 2.729 & 0.114 & 0.428 \\
\cmark &  &  & 2.768 & 0.124 & 0.303 \\
\cmark & \cmark &  & 2.335 & 0.155 & 0.981 \\
 & \cmark & \cmark & 2.336 & 0.180 & 1.081 \\
\cmark &  & \cmark & 1.700 & 0.256 & 2.325 \\
\cmark & \cmark & \cmark & \textbf{1.163} & \textbf{0.386} & \textbf{2.819} \\
\bottomrule
\end{tabular}
}
\label{tbl:ablation_study}
\end{table}

\begin{table}[h]
\centering
\caption{The results of using entire images as input and masking off irrelevant objects on LLMaFF. }
\vspace{2mm}
\resizebox{0.8\textwidth}{!}{ % 修改宽度适应单栏宽度
\begin{tabular}{>{\centering\arraybackslash}p{2.5cm} >{\centering\arraybackslash}p{2.1cm} >{\centering\arraybackslash}p{2.1cm} >{\centering\arraybackslash}p{2.1cm}}
\toprule
Input & KLD$\downarrow$ & SIM$\uparrow$ & NSS$\uparrow$ \\
\midrule
entire image & 2.752 & 0.134 & 0.27 \\
mask off & 1.163  & 0.386 & 2.819 \\
\bottomrule
\end{tabular}
}
\label{tbl:full_images}
\end{table}

\section{Conclusion}
In this paper, we introduce a new task of affordance grounding based on natural language instructions and propose a novel framework, WorldAfford. Our framework uses LLMs to process natural language instructions and employs SAM and CLIP for object segmentation and selection. 
We further propose a Weighted Context Broadcasting module, allowing WorldAfford to localize affordance regions of multiple objects. 
Additionally, we present a new dataset, LLMaFF, to benchmark this task. 
The experimental results demonstrate that WorldAfford outperforms the other state-of-the-art methods for affordance grounding on both the AGD20K dataset and the new LLMaFF dataset.
\clearpage  % TODO REVIEW/FINAL: This \clearpage needs to be removed from both review and camera-ready versions.

\bibliographystyle{splncs04}
\bibliography{main}
\end{document}